\pdfoutput=1
\documentclass[11pt,a4paper]{article}
\usepackage[hyperref]{acl2020}
\usepackage{times}
\usepackage{latexsym}

\usepackage{soul}
\usepackage{xcolor}
\usepackage{listings}
\newcommand{\mt}[1]{{\textcolor{black}{#1}}}
\newcommand{\mn}[1]{{\textcolor{black}{#1}}}

\newcommand{\mg}[1]{{\textcolor{black}{#1}}}

\usepackage{microtype}

\aclfinalcopy 
\def\aclpaperid{31} 

\title{End-to-End Speech-Translation with Knowledge Distillation:\\ FBK@IWSLT2020}

\author{
  Marco Gaido\textsuperscript{1,2},
  Mattia Antonino Di Gangi\textsuperscript{1,2},
  Matteo Negri\textsuperscript{1},
  Marco Turchi\textsuperscript{1} \\
\textsuperscript{1}Fondazione Bruno Kessler, Trento, Italy \\
\textsuperscript{2}University of Trento, Italy \\
\texttt{\{mgaido|digangi|negri|turchi\}@fbk.eu}}

\date{}

\begin{document}
\maketitle
\begin{abstract}
This paper describes FBK's participation in the IWSLT 2020 offline speech translation (ST) task.
The task evaluates systems' ability to translate English TED talks audio into German texts.
The test talks are provided in two versions:
one contains the data already segmented with automatic tools and the other is the raw
data without any segmentation. Participants
can decide whether to work on custom segmentation or not. We used the provided segmentation.    
Our system is an end-to-end model based on an adaptation of the Transformer for speech data.
Its training process is the main focus of this paper and it is based on:
\textit{i)} transfer learning (ASR pretraining and knowledge distillation),
\textit{ii)} data augmentation (SpecAugment, \textit{time stretch} and synthetic data),
\textit{iii)} combining synthetic and real data marked as different domains,
and \textit{iv)} multitask learning using the CTC loss.
Finally, after the training with word-level knowledge distillation is complete, our ST models
are fine-tuned using label smoothed cross entropy.
Our best model scored 29 BLEU on the MuST-C En-De test set, which is an excellent result
compared to recent papers, and 23.7 BLEU on the same
data segmented with VAD, showing the need for researching
solutions addressing this specific data condition.
\end{abstract}
\section{Introduction}
The offline speech translation task consists in generating the text translation
of speech audio recordings into a different language. In particular, the IWSLT2020
task \cite{iwslt_2020} evaluates German translation of English recordings extracted from TED talks.
The test dataset is provided to participants
both segmented in a sentence-like format using a Voice Activity Detector (VAD)
and in the original unsegmented form. Although custom segmentation of the data can provide drastic
improvements in the final scores, in our work we have not addressed it, participating
only with the provided segmentation.

Two main approaches are possible to face the speech translation task. The classic one
is the cascade solution, which includes automatic speech recognition (ASR)
and machine translation (MT) components.
The other option is an end-to-end (E2E) solution, which performs ST
with a single sequence-to-sequence model.
Both of them are allowed for the IWSLT2020 task, but our submission is based on an E2E model.

E2E ST models gained popularity in the last few years. Their rise is due to the
lack of error propagation and the reduced latency in generating the output compared to the
traditional cascaded approach. Despite these appealing properties, they failed so far to reach
the same results obtained by cascade systems, as shown also by last year's  IWSLT campaign \cite{iwsltprec_2019}.
One reason for this is the limited amount of parallel corpora
compared to those used to separately train ASR and MT components.
Moreover, training an E2E ST system is more difficult because the task is more complex,
since it deals with understanding the content of the input audio and translating it
into a different language directly and without recurring to intermediate representations.

The above-mentioned observations have led researchers
to focus on transferring knowledge from MT and ASR systems to improve the ST models.
A traditional approach consists in pretraining components: the ST encoder
is initialized with the ASR encoder and the ST decoder with the MT decoder.
The encoder pretraining has indeed proved to be effective \cite{bansal-etal-2019-pre},
while the decoder pretraining has not demonstrated to be as effective,
unless with the addition of adaptation layers \cite{bahar2019comparative}.
A more promising way to transfer knowledge from an MT model is to use the MT as a teacher
to distill knowledge for the ST training \cite{liu2019endtoend}.
This is the approach we explore in the paper.

Despite its demonstrated effectiveness, ASR pretraining has been replaced
in some works by multitask learning \cite{weiss2017sequencetosequence}.
In this case, the model is jointly trained with two (or more) loss functions
and usually the model is composed of 3 components: \textit{i)} a shared encoder,
\textit{ii)} a decoder which generates the transcription, and \textit{iii)} a decoder which generates
the translation. We adopt the slightly different approach introduced by \cite{bahar2019comparative},
which does not introduce an additional decoder
but relies on the CTC loss in order to predict the transcriptions \cite{kim2016joint}.
As this multi-task learning has been proposed for speech recognition and
has demonstrated to be useful in that scenario,
we also include the CTC loss in ASR pretraining.

Another topic that received considerable attention is data augmentation. Many techniques have been
proposed: in this work we focus on SpecAugment \cite{Park_2019}, \textit{time stretch}
and sub-sequence sampling \cite{nguyen2019improving}.
Moreover, we used synthetic data generated by automatically translating the ASR datasets with our MT model.
This process can also be considered as a sequence-level knowledge distillation technique, named
Sequence KD \cite{kim2016sequencelevel}.

In this paper, we explore different ways to combine synthetic and real data.
We also check if the benefits of the techniques mentioned above are orthogonal and
joining them leads to better results.

Our experiments show that:
\begin{itemize}
    \item knowledge distillation, ASR pretraining, multi-task learning and data augmentation
    are complementary
    \mg{, i.e. }they cooperate to produce a better model;
    \item combining synthetic and real data marking them with different tags \cite{caswell2019tagged}
    leads to a model which generalizes better;
    \item fine-tuning a model trained with word-level knowledge distillation
    using the more classical label smoothed cross entropy \cite{szegedy2016rethinking}
    significantly improves the results;
    \item there is a huge performance gap between data segmented in sentences and data segmented with VAD.
    Indeed, on the same test set, the score on VAD-segmented data is lower by 5.5 BLEU.
\end{itemize}

\mg{\mt{To summarize}, our submission is characterized
by tagged synthetic data, multi-task with CTC loss on the transcriptions,
data augmentation and word-level knowledge distillation.}

\section{Training data}
This section describes the data used to build our models. They include:
\textit{i)} MT corpora (English-German sentence pairs), for the model used in knowledge distillation;
\textit{ii)} ASR corpora (audio and English transcriptions),
for generating a pretrained encoder for the ST task;
\textit{iii)} ST corpora (audios with corresponding English transcription and German translation),
for the training of our ST models.
For each task, we used all the relevant datasets allowed by the evaluation
campaign\footnote{\label{task_webpage}\url{http://iwslt.org/doku.php?id=offline\_speech\_translation}}.

\paragraph{MT.}
All datasets allowed in WMT 2019 \cite{barrault-etal-2019-findings} were used for the MT training, with
the addition of OpenSubtitles2018 \cite{Lison2016OpenSubtitles2016EL}.
These datasets contain spurious sentence pairs: some target sentences are
in a language different from German (often in English) or are unrelated
to the corresponding English source or contain unexpected characters (such as ideograms).
As a consequence,
an initial training on them caused the model to produce some
English sentences, instead of German, in the output.
Hence, we cleaned our MT training data using Modern MT \cite{modernmt}\footnote{\mg{We \mt{run} the \lstinline{CleaningPipelineMain} class of MMT.}},
in order to remove sentences whose language is not the correct one.
We further filtered out sentences containing ideograms with a custom script.
Overall, we removed roughly 25\% of the data and
the final dataset used in the training contains nearly 49 million sentence pairs.

\paragraph{ASR.}
For this task, we used both pure ASR and ST available corpora. They include TED-LIUM 3 \cite{Hernandez_2018},
Librispeech \cite{librispeech}, Mozilla Common Voice\footnote{\url{https://voice.mozilla.org/}},
How2 \cite{sanabria18how2}, the En-De section of MuST-C \cite{mustc}, the  Speech-Translation TED corpus
provided by the task organizers\textsuperscript{\ref{task_webpage}} and
the En-De section of Europarl-ST \cite{europarlst}. All data was lowercased and punctuation was removed.

\paragraph{ST.}
In addition to the allowed ST corpora (MuST-C, Europarl-ST and the Speech-Translation TED corpus),
we generated synthetic data using Sequence KD (see Section \ref{data_augm})
for all the ASR datasets missing the German reference.
Moreover, we generated synthetic data for the
En-Fr section of MuST-C. Overall, the combination of real and generated data
resulted in a ST training set of 1.5 million samples.

All texts were preprocessed by tokenizing them, de-escaping special characters and normalizing
punctuation with the scripts in the Moses toolkit  \cite{koehn-etal-2007-moses}.
The words in both languages were segmented using BPE with 8,000 merge rules
learned jointly on the two languages of the MT training data \cite{sennrich2015neural}.
The audio was converted into 40 log Mel-filter banks with speaker normalization using
XNMT \cite{neubig-etal-2018-xnmt}.
We discarded samples with more than 2,000 filter-banks in order to prevent memory issues.

\section{Models and training}

\subsection{Architectures}
The models we trained are based on Transformer \cite{transformer}. The MT model is a plain Transformer with 6 layers for
both the encoder and the decoder, 16 attention heads, 1,024 features for the attention layers and
4,096 hidden units in feed-forward layers.

\begin{table}[ht]
\small
\centering
\begin{tabular}{rrr|r}
2D Self-Attention & Encoder & Decoder & BLEU  \\
\hline
2 & 6 & 6 & 16.50 \\
0 & 8 & 6 & \textbf{16.90} \\
\hline \hline
2 & 9 & 6 & 17.08 \\
2 & 9 & 4 & 17.06 \\
2& 12 & 4 & \textbf{17.31} \\
\end{tabular}
\caption{Results on Librispeech with Word KD varying the number of layers.}
\label{tab:libri_layers}
\end{table}

The ASR and ST models are a revisited version of the S-Transformer introduced by
\cite{digangi2019adapting}. In preliminary experiments on Librispeech (see Table \ref{tab:libri_layers}),
we observed that replacing 2D self-attention layers with additional Transformer
encoder layers was beneficial to the final score.
Moreover, we noticed that adding more layers in the encoder improves the results, while removing
few layers of the decoder does not harm performance.
Hence, the models used in this work process the input with two 2D CNNs,
whose output is projected into the higher-dimensional space used by the Transformer encoder layers.
The projected output is summed with positional embeddings
before being fed to the Transformer encoder layers, which use logarithmic distance penalty.

Both our ASR and ST models have 8 attention heads, 512 features for the attention layers
and 2,048 hidden units in FFN layers.
The ASR model has 8 encoder layers and 6 decoder layers, while the ST model has
11 encoder layers and 4 decoder layers. The ST encoder is initialized with the ASR encoder
(except for the additional 3 layers that are initialized with random values). The decision of having
a different number of encoder layers in the two encoders is motivated by the idea of introducing adaptation layers,
which \cite{bahar2019comparative} reported to be essential when initializing the decoder
with that of a pretrained MT model.

\subsection{Data augmentation}
\label{data_augm}

One of the main problems for end-to-end ST is the scarcity of parallel corpora.
In order to mitigate this issue, we explored the following
data augmentation strategies in our participation.

\textbf{SpecAugment.}
SpecAugment is a data augmentation technique originally introduced for ASR, whose
effectiveness has also been demonstrated for ST \cite{bahar2019using}.
It operates on the input filterbanks and it consists in masking consecutive portions
of the input both in the frequency and in the time dimensions. On every input, at each
iteration, SpecAugment is applied with probability $p$. In case of application,
it generates \textit{frequency masking num} masks on the frequency axis and \textit{time masking num}
masks on the time axis. Each mask has a starting index,
which is sampled from a uniform distribution,
and a number of consecutive items to mask, which is a random number between 0 and respectively
 \textit{frequency masking pars} and \textit{time masking pars}. Masked items are set to 0.
In our work, we \mg{always} applied SpecAugment to both the ASR pretraining and the ST training.
The configuration we used are: $p$ = 0.5, \textit{frequency masking pars} = 13,
\textit{time masking pars} = 20, \textit{frequency masking num} = 2 and \textit{time masking num} = 2.

\textbf{Time stretch.}
\textit{Time stretch} \cite{nguyen2019improving} is another technique which operates directly
on the filterbanks, aiming at generating the same effect of speed perturbation.
It divides the input sequence in windows of \textit{w} features and re-samples each of them
by a random factor \textit{s} drawn by a uniform distribution between 0.8 and 1.25 (in our implementation,
the lower bound is set to 1.0 in case of an input sequence with length lower than 10).
In this work, we perturb an input sample using \textit{time stretch} with probability 0.3.

\textbf{Sub-sequence sampling.}
As mentioned in the introduction, there is a huge gap in model's performance when translating
data split in well-formed sentences and data split with VAD. In order to reduce this difference,
we tried to train the model on sentences which are not always well-formed by using sub-sequence sampling \cite{nguyen2019improving}.
Sub-sequence sampling requires the alignments between the speech and the target text at word level.
As this information is not possible to obtain for the translations, we created the sub-sequences
with the alignments between the audio and the transcription, and then we translated the obtained
transcription with our MT model to get the target German translation.
For every input sentence, we generated three segments:
\textit{i)} 
one starting at the beginning of the sentence and ending at a random word
in the second half of the sentence,
\textit{ii)} one starting at a random word
in the first half of the sentence and ending at the end of the sentence,
and \textit{iii)} one starting at a random word in the first quarter of the sentence and ending at a random
word in the last quarter of the sentence.

In our experiments, this technique has not provided significant improvements (the gain was less
than 0.1 BLEU on the VAD-segmented test set). Hence, it was not included in our final models.

\textbf{Synthetic data.}
Finally, we generated synthetic translations for the data in the ASR datasets to
create parallel audio-translation pairs to be included in the ST trainings.
The missing target sentences were produced by translating the transcript of each audio sample with our MT model,
as in \cite{jia2018leveraging}.
If the transcription of a dataset was provided with punctuation and correct casing,
this was fed to the MT model; otherwise, we had to use the lowercase transcription without punctuation.

\begin{table}[ht]
\small
\centering
\begin{tabular}{r|r}
Top K & BLEU  \\
\hline
4 & 16.43 \\
8 & \textbf{16.50} \\
64 & 16.37 \\
1024 & 16.34 \\
\end{tabular}
\caption{Results on Librispeech with different K values, where K is the number of tokens considered for Word KD.}
\label{tab:libri_topk}
\end{table}

\subsection{Knowledge distillation}
While the ASR and MT models are optimized on label smoothed cross entropy with smoothing factor
0.1, our ST models are trained with word-level knowledge distillation (Word KD).
In Word KD, the model being trained is named \textit{student} and the goal is to teach it to produce
the same output distribution of another - pretrained - model, named \textit{teacher}.
This is obtained by computing the KL divergence \cite{kullback1951} between the distribution produced by the
student and the distribution produced by the teacher. The rationale of knowledge distillation resides
in providing additional information to the student, as the output probabilities produced by the teacher
reflect its hidden knowledge (the so-called \textit{dark knowledge}), and in the fact that the soft
labels produced by the teacher are an easier target to match for the student than cross entropy.

In this work, we follow \cite{liu2019endtoend}, so the teacher model is our MT model and the student is the ST model.
Compared to \cite{liu2019endtoend}, we make the training more efficient
by extracting only the top 8 tokens from the teacher distribution.
In this way, we can precompute and store the MT output instead of computing it at each training iteration,
since its size is reduced by three orders of magnitude.
Moreover, this approach does not affect negatively the final score, as shown by \cite{tan2018multilingual}
and confirmed for ST by our experiments in Table \ref{tab:libri_topk}).

Moreover, once the training with Word KD is terminated, we perform a fine-tuning of
the ST model using the label smoothed cross entropy. Fine-tuning on a different target is
an approach whose effectiveness has been shown by \cite{kim2016sequencelevel}.
Nevertheless, they applied a fine-tuning on knowledge distillation after a
pretraining with the cross entropy loss, while here we do the opposite. Preliminary
experiments on Librispeech showed that there is no difference in the order of the
trainings (16.79 vs 16.81 BLEU, compared to 16.5 BLEU before the fine-tuning).
In the fine-tuning, we train both on real and synthetic data, but we do not
use the other data augmentation techniques.

\subsection{Training scheme}
A key aspect is the training scheme used to combine the real and synthetic datasets.
In this paper, we explore two alternatives:

\begin{itemize}
    \item \textbf{Sequence KD + Finetune}:
    this is the training scheme suggested in \cite{he2019revisiting}.
    The model is first trained with Sequence KD and Word KD on the synthetic datasets and
    then it is fine-tuned on the datasets with ground-truth targets using Word KD.
    \item \textbf{Multi-domain}: similarly to our last year submission \cite{iwslt_fbk_2019},
    the training is executed on all data at once, but
    we introduce three \textit{tokens} representing the three types of data, namely:
    \textit{i)} those whose ground-truth translations are provided,
    \textit{ii)} those generated from true case transcriptions with punctuation,
    \textit{iii)} those generated from lowercase transcriptions without punctuation.
    We explore the two most promising approaches according to \cite{gangi2019onetomany}
    to integrate the token with the data, i.e. 
    summing the token to all input data and summing the token to all decoder input embeddings.
\end{itemize}

\subsection{Multi-task training}
We found that adding the CTC loss \cite{Graves2006ConnectionistTC} to the training objective
gives better results both in ASR and ST, although it slows down the training by nearly a factor of 2.
During the ASR training, we added the CTC loss on the output of the last layer of the encoder.
During the ST training, instead, the CTC loss was computed using the output of the last
layer pretrained with the ASR encoder, ie. the 8th layer. In this way, the ST encoder
has three additional layers which can transform the representation into features which are
more convenient for the ST task, as \newcite{bahar2019comparative} did
introducing an adaptation layer.

\section{Experimental settings}
For our experiments, we used the described training sets and we picked the best model
according to the perplexity on MuST-C En-De validation set.
We evaluated our models on three benchmarks:
\textit{i)} the MuST-C En-De test set segmented at sentence level,
\textit{ii)} the same test set segmented with a VAD \cite{meigner2010lium},
and \textit{iii)} the IWSLT 2015 test set \cite{Cettolo2015TheI2}. 

We trained with Adam \cite{adam} (betas \textit{(0.9, 0.98)}). Unless stated otherwise,
the learning rate was set to increase linearly from 3e-4 to 5e-4 in the first 5,000 steps
and then decay with an inverse square root policy. For fine-tuning, the learning rate
was kept fixed at 1e-4. A 0.1 dropout was applied.

Each GPU processed mini-batches containing up to 12K tokens or 8 samples and updates were performed every 8
mini-batches. As we had 8 GPUs, the actual batch size was about 512.
In the case of multi-domain training, a batch for each domain was processed before an update:
since we have three domains, the overall batch size was about 1,536.
Moreover, the datasets in the different domains had different sizes,
so the smaller ones were oversampled to match the size of the largest.

As the truncation of the output values of the teacher model to the top 8 leads to a more peaked
distribution, we checked if contrasting this bias is beneficial or not.
Hence, we tuned the value of the temperature at generation time in the interval 0.8-1.5.
The temperature $T$ is a parameter which is used to divide the $logits$ before the $softmax$
and determines whether to output
a softer (if $T > 1$) or a sharper (if $T < 1$) distribution \cite{hinton2015distilling}.
By default $T$ is 1, returning an unmodified distribution.
The generation of the results reported in this paper was performed using $T$ = 1.3 for the models trained
on Word KD. This usually provided a 0.1-0.5 BLEU increase on our
benchmarks compared to $T$ = 1, confirming our hypothesis that a compensation of the bias
towards a sharper distribution is useful.
Instead, the $T$ was set to 1 during the generation with models
trained with label smoothed cross entropy, as in this case a higher (or lower) temperature
caused performance losses up to 1 BLEU point.

All experiments were executed on a single machine with 8 Tesla K80 with 11GB RAM.
Our implementation is built on top of fairseq \cite{ott-etal-2019-fairseq},
an open source tool based on PyTorch \cite{paszke2017automatic}.

\begin{table}[ht]
\small
\begin{tabular}{p{3cm}|p{1.1cm}|p{1.1cm}|p{1cm}}
Model & MuST-C sentence & MuST-C VAD & IWSLT 2015 \\
\hline
Seq KD+FT (w/o TS) & 25.80 & 20.94 & 17.18 \\
\hspace{3mm} + FT w/o KD & 27.55 & 19.64 & 16.93 \\
Multi ENC (w/o TS) & 25.79 & 21.37 & 19.07 \\
\hspace{3mm} + FT w/o KD & 27.24 & 20.87 & 19.08 \\
\hline \hline
Multi ENC+DEC PT & 25.30 & 20.80 & 16.76 \\
\hspace{3mm} + FT w/o KD & 27.40 & 21.90 & 18.55 \\
\hline \hline
Multi ENC+CTC & \textit{27.06} & \textit{21.58} & \textit{20.23} \\
\hspace{3mm} + FT w/o KD (1) & 27.98 & 22.51 & 20.58 \\
Multi ENC+CTC (5e-3) & 25.44 & 20.41 & 16.36 \\
\hspace{3mm} + FT w/o KD & \textbf{29.08} & \textbf{23.70} & 20.83 \\
\hspace{3mm} + AVG 5 (2) & 28.82 & 23.66 & \textbf{21.42} \\
\hline \hline
Multi DEC+CTC (5e-3) & 26.10 & 19.94 & 17.92 \\
\hspace{3mm} + FT w/o KD & 28.22 & 22.61 & 18.31 \\
\hline \hline
Ensemble (1) and (2) & \textbf{29.18} & \textbf{23.77} & \textbf{21.83} \\
\end{tabular}
\caption{Case sensitive BLEU scores for our E2E ST models.
Notes: Seq KD: Sequence KD;
FT: finetuning on ground-truth datasets;
TS: time stretch;
Multi ENC: multi-domain model with sum of the language token to the encoder input;
Multi DEC: multi-domain model with sum of the language token to the decoder input;
DEC PT: pretraining of the decoder with that of an MT model;
CTC: multitask training with CTC loss on the 8th encoder layer in addition to the target loss;
FT w/o KD: finetuning on all data with label smoothed cross entropy;
5e-3: indicates the learning rate used;
AVG 5: average 5 checkpoints around the best.}
\label{tab:results}
\end{table}

\section{Results}

The MT model used as teacher for Sequence KD and Word KD scored 32.09 BLEU on the MuST-C En-De test set.
We trained also a smaller MT model to initialize the ST decoder with it.
Moreover, we trained two ASR models. One without the multitask CTC loss and one with it.
They scored respectively 14.67 and 10.21 WER.
All the ST systems having CTC loss were initialized with the latter,
while the others were initialized with the former.

Table \ref{tab:results} shows our ST models' results computed on the MuST-C En-De and IWSLT2015 test set.

\subsection{Sequence KD + Finetune VS Multi-domain}
First, we compare the two training schemes examined. As shown in Table \ref{tab:results},
\textit{Sequence KD + Finetune} [Seq KD+FT] has the same performance
as \textit{Multi-domain} with language token summed to the input [Multi ENC] 
(or even slightly better) on the MuST-C test set, but it is
significantly worse on the two test set segmented with VAD.
This can be explained by the higher generalization capability
of the \textit{Multi-domain} model. Indeed, \textit{Sequence KD + Finetune} seems to overfit
more the training data; thus, on data coming from a different distribution, as VAD-segmented data are,
its performance drops significantly. 
For this reason, all the following experiments use the \textit{Multi-domain} training scheme.

\subsection{Decoder pretraining and \textit{time stretch}}
The pretraining of the decoder with that of an MT model does not bring
consistent and significant improvements across the test sets [Multi ENC+DEC PT].
Before the fine-tuning with label smoothed cross entropy, indeed,
the model performs worse on all test sets. The fine-tuning, though,
helps improving performances on all test sets,
which was not the case with the previous training.
This can be related to the introduction of \textit{time stretch},
which reduces the overfitting to the training data.
Therefore, we decided to discard the MT pretraining and keep \textit{time stretch}.

\subsection{CTC loss and learning rate}
The multitask training with CTC loss, instead, improves
the results consistently. The model trained with it [Multi ENC+CTC]
outperforms all the others on all test sets by up to 1.5 BLEU points.
During the fine-tuning of these models, we do not perform multitask
training with the CTC loss, so the fine-tuning training is exactly
the same as for previous models.

Interestingly, increasing the learning rate [Multi ENC+CTC (5e-3)], the performance before
the fine-tuning is worse, but the fine-tuning of this models brings
an impressive improvement over all test sets. The reason of this
behavior is probably related to a better initial exploration of
the solution space thanks to the higher learning rate, which,
on the other side, prevents to get very close to the local optimum found.
In this scenario, the fine-tuning with a lower learning rate helps
getting closer to the local optimum, in addition to the usual
benefits.

\subsection{Token integration strategy}
Finally, we tried adding the language token to the embeddings
provided to the decoder, instead of the input data [Multi DEC+CTC (5e-3)].
This was motivated by the idea that propagating this information through
the decoder may be more difficult due to the CTC loss, which is
not dependent on that information so it may hide it to higher layers.
The experiments disproved this hypothesis, as after the fine-tuning
the results are lower on all benchmarks.

\subsection{Submissions}
We averaged our best model over 5 checkpoints, centered in the best
according to the validation loss.
We also created an ensemble with the resulting model and the best among the others.
Both operations were not useful on the two variants of the MuST-C test set, but improved
the score on the IWSLT2015 test set. We argue this means that they are
more robust and generalize better.

\mg{Our \textit{primary} submission has been 
\mn{obtained}
with the ensemble of two models, scoring 20.75 BLEU on the 2020 test set and 19.52 BLEU on the 2019 test set.
Our \textit{contrastive} submission has been generated with the 5 checkpoints average of our best model,
scoring 20.25 BLEU on the 2020 test set and 18.92 BLEU on the 2019 test set.}

\section{Conclusions}
We described FBK's participation in IWSLT2020 offline speech translation
evaluation campaign \cite{iwslt_2020}. Our work focused on the integration of transfer learning,
data augmentation, multi-task training and the training scheme used to
combine real and synthetic data.
Based on the results of our experiments, our submission is characterized
by a multi-domain training scheme, with additional CTC loss on the transcriptions
and word-level knowledge distillation, followed by a fine-tuning on
label smoothed cross entropy.

Overall, the paper demonstrates that the combination of the above-mentioned
techniques can improve the performance of end-to-end ST models so that
they can be competitive with cascaded solutions.
\mg{Moreover, it shows that \mn{\textit{i)}} tagged synthetic data
leads to more robust models than a pretraining on synthetic data followed by a fine-tuning on datasets \mn{with} ground-truth targets
and 
\mn{\textit{ii)}}
fine-tuning on label smoothed cross entropy after a training
with knowledge distillation brings significant improvements.
The}
huge gap (5.5 BLEU) between data segmented in sentences and
data segmented with VAD highlights the need of custom solutions for the latter.
In light of these considerations, our future research will focus on
techniques to improve the results when the audio segmentation is
challenging for ST models.

\section*{Acknowledgments}
This work is part of the ``End-to-end Spoken Language Translation in Rich Data Conditions'' project,\footnote{\url{https://ict.fbk.eu/units-hlt-mt-e2eslt/}} which is financially supported by an Amazon AWS ML Grant.

\bibliography{acl2020}
\bibliographystyle{acl_natbib}

\end{document}